# CONCURRENT LEXICALIZED DEPENDENCY PARSING: A BEHAVIORAL VIEW ON *ParseTalk* EVENTS

Susanne Schacht, Udo Hahn & Norbert Bröker

𝒞ℒℐℱ - Computational Linguistics Research Group
Freiburg University
D-79085 Freiburg, Germany

email: {sue, hahn, nobi}@coling.uni-freiburg.de

*Abstract.* The behavioral specification of an object-oriented grammar model is considered. The model is based on full lexicalization, head-orientation via valency constraints and dependency relations, inheritance as a means for non-redundant lexicon specification, and concurrency of computation. The computation model relies upon the actor paradigm, with concurrency entering through asynchronous message passing between actors. In particular, we here elaborate on principles of how the global behavior of a lexically distributed grammar and its corresponding parser can be specified in terms of event type networks and event networks, resp.

## 1 INTRODUCTION

In this paper, we propose a grammar model that combines lexical organization of grammatical knowledge with lexicalized control of the corresponding parser in a coherent object-oriented specification framework. We build upon recent developments in the field of linguistic grammar theory which have already yielded a rigid *lexical modularization*, but extend them by assigning full procedural autonomy to lexical units. In particular, we treat lexical items as active *lexical processes* communicating with each other by message passing. Thus, they dynamically establish heterogeneous communication lines in order to determine each lexical item's functional role. While the issue of lexicalized control has early been investigated in the paradigm of conceptual parsers (e.g., Riesbeck & Schank 1978), and word expert parsing in particular (Small & Rieger, 1982), we here elaborate on improving its lexical communication facilities by formalizing the parser's *message passing* protocol according to actor computation principles. As this protocol allows for asynchronous message passing, *concurrency* enters as a theoretical notion at the level of grammar specification, not only as an implementational feature. Correspondingly, we introduce a behavioral description in terms of *event type networks* which represent grammatical interrelations at the level of actor definitions, while *event networks* represent the parsing process in terms of actual messages exchanged between instantiated actors. The *ParseTalk* model outlined in this paper can therefore be considered as an attempt to remedy the lack of theoretical integration of parallelism at the level of grammar design.

## 2 *ParseTalk*'s GRAMMAR MODEL

The *ParseTalk* grammar model (cf. Bröker, Hahn & Schacht (1994) for a more comprehensive treatment) considers dependency relations between words as the fundamental notion of linguistic analysis. This corresponds to the head-orientation found in most modern grammar theories. Grammatical specifications are given in the format of *valency constraints* attached to each lexical unit, on which the computation of concrete *dependency relations* is based. A modifier is said to depend on its head if the modifier satisfies the constraints placed on it. These constraints incorporate information about the hierarchy of word classes (encapsulating declarative and behavioral properties of lexical items), morphosyntax (containing the grammatical conditions of the combination of lexical items to phrases as expressed by a unification formalism, similar to Shieber, 1986), linear ordering (stating precedence relations between a head and its modifiers), and permitted conceptual roles (expressed in terms of a hybrid, classification-based knowledge representation formalism; cf. MacGregor, 1991). *Dependencies* are thus asymmetric binary relations that can be established by local computations involving only two lexical items[1] and simultaneously take grammatical as well as conceptual well-formedness criteria into account.

By way of *inheritance* (for a recent survey of applying inheritance mechanisms in modern grammar theory, cf. Daelemans, De Smedt & Gazdar, 1992) the entire collection of lexical items is organized in a *lexical hierarch*y, the lexical items forming its leaves and the intermediary nodes representing grammatical generalizations in terms of word classes. This form of specification is similar to various proposals currently investigated within the unification grammar community (e.g., Evans & Gazdar, 1990).

---

[1] We extend this definition to incorporate the notion of phrases as well. Although phrases are not explicitly represented (e.g., by non-lexical categories), we consider each complete subtree of the dependency tree a phrase (this definition allows discontinuous phrases as well). A dependency is thus not treated as a relation between words (as in Word Grammar (Hudson, 1990, p.117), but between a word and a dependent phrase (as in Dependency Unification Grammar (Hellwig, 1988)). The root of a phrase is taken to be the representative of the whole phrase.

# 3 *ParseTalk*'s COMPUTATION MODEL

Although the object-oriented paradigm seems to be well suited to support the distribution of data through encapsulation and the distribution of control via message passing, most object-based calculi rely on synchronous messages and therefore do not provide for concurrency. One of the few exceptions that aim at the methodologically clean combination of object-oriented features with concurrency and distribution is the actor model of computation (Agha & Hewitt, 1987). It assumes a collection of independent objects, the *actors*, communicating via asynchronous, point-to-point message passing. All messages are guaranteed to be delivered and processed, but in an unpredictable order and indeterminate time. Each actor has an *identity* (its mail address), a *state* (consisting of the addresses of its *acquaintances*, i.e., the set of other actors it may send messages to) and a *behavior* (i.e., its reaction to incoming messages). The arrival of a message at an actor is called an *event*; it triggers an action described by the corresponding *method definition*, a composition of the following atomic actions: creating a new actor (<u>create</u> actorType (acquaintances)); sending a message to an acquainted or a newly created actor (<u>send</u> actor message); or specifying new acquaintances for itself (<u>become</u> (acquaintances)). An actor system is dynamic, since new actors can be created and the communication topology is reconfigurable in the course of actor computations.

The actor model does not contain synchronization primitives, but we assume one-at-a-time serialized actors for our specification, i.e., actors that cannot process more than one message at a time and that process each message step by step (cf. Hewitt & Atkinson (1979) for expressing this convention in terms of patterns of simple actors). The distribution of computation among the collection of actors is thus the only source of parallelism. Furthermore, in order to compute complex, but well understood and locally determined linguistic conditions and functions, such as unification of feature structures and queries sent to a (conceptual) knowledge base, we establish a synchronous request-reply protocol (cf. Lieberman, 1987).

The *ParseTalk* model extends the formal foundations of the basic actor model according to the requirements set up by the natural language processing application. These extensions are expressible by the primitives of the basic model. We distinguish between *word actors*, *relations* between word actors and a special set of *messages* word actors exchange.

- **Word Actors:** The grammatical knowledge associated with each lexical item is represented in a *word actor definition*. Upon instantiation of a specific word actor, the acquaintances specified in the definition will be initialized with actors which stand for the lexical item's morphosyntactic features, its conceptual representation, valency constraints and, after instantiation and subsequent parsing events, governed lexical items and further grammatical relations (e.g., adjacency, textual relations).

- **Word actor relations:** Acquaintances of word actors are *tagged* according to linguistic criteria in order to serve as navigation aids in linguistic structures (the message distribution mechanism described below). Textual relations, e.g., are distinguished from linear adjacency and hierarchical dependency relations. Tagging imposes a kind of typing onto acquaintances that is missing in other actor systems.

- **Word actor messages:** In contrast to simple messages which unconditionally trigger the execution of the corresponding method at the receiving actor, we define *complex word actor messages* as full-fledged actors with independent computational capabilities. Departure and arrival of complex messages are actions which are performed by the message itself, taking the sender and the target actors as parameters. Upon arrival, a complex message determines whether a copy is forwarded to selected acquaintances of its receiver and whether the receiver may process the message on its own. Hence, we redefine an arrival event to be an uninterruptable sequence of a computation event <u>and</u> distribution events. The *computation event* corresponds to an arrival of a simple message at the receiving word actor, i.e. an event in the basic model; it consists of the execution of an actor's method that may change the actor's state and trigger additional messages. The *distribution events* provide for the forwarding of the message and are realized by creating new complex messages. They depend on the (unchanged) state of the receiving actor or on the result of the computation event and take place before and after the computation event. This extension accounts for the complexity of interactions between word actors.

We define the semantics of an actor program in terms of two kinds of networks. First, we consider *event types* which refer to message keys and can easily be determined from a given actor program. Next, we turn to actual events that involve instantiated actors. Both, event types and events, are partially ordered by the transitive closures of relations among them, *causes$^t$* and *causes*, resp., that give rise to *event type network*s and *event network*s.

A *program* (in our application: a lexical grammar) is given by a set of *actor definitions*. The definition characterizes the type of an actor. Given a program, *event types*, written as $[* \Leftarrow key]$, can be syntactically determined by inspecting the method definitions within the program. Let an actor type aName be defined by:

<u>defActor</u> aName (acquaintance$_1$ ... acquaintance$_k$)
    <u>meth</u> key$_1$(param$_1$ ... param$_m$) (action$_1$)
    ...
    <u>meth</u> key$_n$ (param$_1$ ... param$_l$) (action$_n$)

with action$_i$ defined by the following grammar fragment:

action    ::= action; action
       | <u>if</u> condition (action) [ <u>else</u> (action) ]
       | <u>send</u> actor messageKey ( param$^*$ )
       | <u>become</u> ( acquaintance$^*$ )

We may now map message keys to sets of message keys, defining the function *script*$_{aName}$ as follows:

$script_{aName}: Keys \to 2^{Keys}$.

$script_{aName}(key_i) = send(action_i)$ with

$send(action) :=$
$\begin{cases} \{msgKey\} & \text{if action} = \underline{send} \text{ actor msgKey (param, ...)} \\ send(a_1) \cup send(a_2) & \text{if action} = \underline{if} \text{ condition } a_1 \underline{else} \, a_2 \\ send(a_1) & \text{if action} = \underline{if} \text{ condition } a_1 \\ send(a_1) \cup send(a_2) & \text{if action} = a1; a2 \\ \varnothing & \text{else} \end{cases}$

For a program P, *script* is the union of all given $script_{name}$ with $name \in \{ aName \mid P$ contains a definition for $aName \}$ and yields a set containing the keys of those messages that can be provoked by a message with the key mKey. Now, a relation between event types is defined by $causes^t$:

$([* \Leftarrow mKey], [* \Leftarrow nKey]) \in causes^t$

$:\Leftrightarrow nKey \in script(mKey)$.

Turning to actual events now, we define an actor $\chi$ as being composed of an *identity* n (taken from the set of natural numbers, **N**), a *state* $\in \mathcal{S}$ and a *behavior* $\in \mathcal{B}$. Hence, $\mathcal{A}$, the set of actors, is a subset of $\mathbf{N} \times \mathcal{S} \times \mathcal{B}$.

$\mathcal{S} = 2^{\{(y: z) \mid y \text{ is an identifier}, z \in \mathcal{A}\}}$, an element of $\mathcal{S}$ associates acquaintance names and values, which are actors. Since actors change their acquaintances, their state is valid *in time*, i.e. at a particular event. The state of an actor $a$ receiving a message $m$ will be written as $s_{a,[a \Leftarrow m]}$. State changes caused by the message apply at the end of the event $[a \Leftarrow m]$ (by executing a <u>become</u> action).

$\mathcal{B}$ is a set of functions, defined as follows: The state $s_{\chi,e}$ of an actor $\chi$ at the event $e$ (the reception of a message m) is determined by its initial state given after its creation event, and the repeated application of its state transition function, $transit_{\chi}$, which maps pairs of states ($s \in \mathcal{S}$) and messages (m $\in \mathcal{M} \subset \mathcal{A}$) to new states:

$transit_{\chi}: (\mathcal{S} \times \mathcal{M}) \to \mathcal{S}$

The <u>send</u> actions an actor $\chi$ performs at a particular event are expressed as pairs of target actors and messages to be sent; the target actors are either acquaintances of the sending actor or supplied as message parameters. They are determined by the function

$task_{\chi}: (\mathcal{S} \times \mathcal{M}) \to 2^{(\pi_1(\mathcal{A}) \times \mathcal{M})}$

where $\pi_1(\mathcal{A})$ denotes the projection onto the first component of $\mathcal{A}$, viz. **N**.

The *behavior* of an actor $\chi$ can then be stated by the function $behave_{\chi} \in \mathcal{B}$ that combines $transit_{\chi}$ and $task_{\chi}$ in that it maps pairs of states and messages to pairs consisting of the new state of the actor and a set of pairs of target actor identities and messages, viz.,

$behave_{\chi}: (\mathcal{S} \times \mathcal{M}) \to (\mathcal{S} \times 2^{(\pi_1(\mathcal{A}) \times \mathcal{M})})$.

Abstracting from a local actor perspective the behavior of an entire actor system (in our application: the lexical parser composed of a collection of word actors) is determined by the way multiple events are related under the *causes* relation (though events are written as [actor $\Leftarrow$ message], the message key is used as an abbreviation for the messages in Section 5):

$([a \Leftarrow m], [b \Leftarrow n]) \in causes$

$:\Leftrightarrow (\pi_1(b), n) \in task_a(s_{a,[a \Leftarrow m]}, m)$.

Events that are not ordered by the transitive closure of *causes* can take place in parallel or, if they refer to the same actor, in an arbitrary order.

## 4 EVENT TYPE NETWORK SPECIFICATION OF A GRAMMAR FRAGMENT

The protocol (messages and associated actions) for establishing dependencies outlined below encodes structural restrictions of the dependency structure (projectivity), ensures incremental generation of dependency trees, and provides a domesticated form of concurrency.

Consider a newly instantiated word actor $w_n$ (cf. Fig.1) searching bottom-up for its head by sending a **searchHead** message to its immediate left neighbor, $w_{n-1}$. The **searchHead** message is recursively forwarded by a sequence of distribution events to the head of each receiving actor (i.e., $w_{n-1}$, $w_k$, $w_j$); messages only pass the outer fringe of the already established dependency tree (these are circled in Fig.1). Since only the actors receiving the **searchHead** message may later govern $w_n$, projective trees are generated[2].

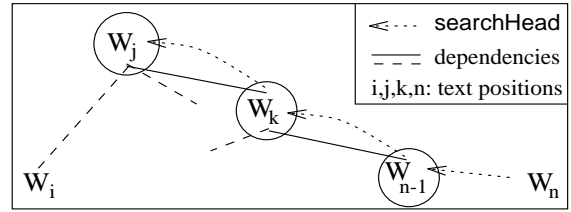

**Figure 1.** Forwarding a search message

To allow for domesticated concurrency as required for adequate linguistic and cognitive processing (Clark & Gibson, 1988), a receipt protocol allows $w_n$ to keep track of all events (transitively) caused by its **searchHead** message. This protocol requires each actor receiving a **searchHead** message to reply to the initiator of the **searchHead** message by a **receipt** message when the receivers computation has finished[3]. Since complex messages can be quasi-recursively forwarded, the number of replies cannot be determined in advance. Therefore, the **receipt** message contains all actors to which the **searchHead** message has been distributed, enabling the initiator $w_n$ to keep track of all receivers and wait for a **receipt** message from each[4]. Only after all events caused by the **searchHead** message have terminated, the next word actor $w_{n+1}$ is instantiated by sending a **scanNext** message to the text scanning actor.

---

[2] Of course, $w_n$ may be governed by any word actor governing $w_j$, but due to the incrementality of the analysis, each head of $w_j$ must be located to the right of $w_n$.
[3] Note that "computation" here may include a number of events that are caused by the **searchHead** message, viz. the **headFound** and **headAccepted** messages described below.
[4] We plan to extend our algorithm to a generic termination detection scheme similar to the proposal in Shavit & Francez, 1986.

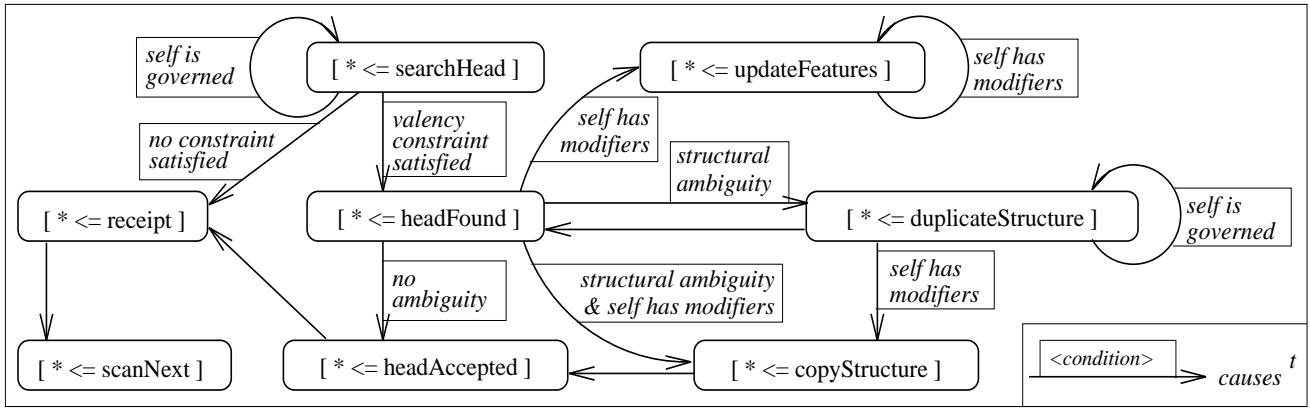

**Figure 2.** Event type network

Upon reception of a **searchHead** message, a word actor $w_k$ checks whether $w_n$ satisfies the constraints for one of $w_k$'s valencies. If no constraints are satisfied, a **receipt** message is sent back to signal termination of this particular event at $w_k$. If $w_n$ may fill a valency of $w_k$, a **headFound** message is sent back to $w_n$, thus possibly imposing additional grammatical restrictions on the targeted item. If $w_n$ is still ungoverned, it adjusts its grammatical description (and those of its modifiers, if necessary, by sending **updateFeatures** to each) and signals acceptance of the new head by a **headAccepted** message directed to $w_k$. These interrelations are summarized in the event type network in Fig.2.

This three-step protocol allows alternative attachments to be checked in parallel (concurrent processing of **searchHead** messages at different actors). Structural ambiguities are detected whenever a **headFound** message arrives at an actor $w_n$ which is already governed. In this case, $w_n$ duplicates itself and its modifiers (using the **copyStructure** message), resulting in $\overline{w}_n$, and asks the prospective head to copy itself (by sending a **duplicateStructure** message). $\overline{w}_n$ becomes head of the copies of the modifiers of $w_n$ (because each is answering the **copyStructure** message with a **headAccepted** message) and will be governed by the copy of the head (because the copy sends another **headFound** message to $\overline{w}_n$; for a more detailed discussion, cf. Hahn, Schacht & Bröker, forthcoming).

The unpacked representation of ambiguity is necessary because of the simultaneous incorporation of conceptual analysis into the parsing process. Different syntactic structures result in different conceptual analyses, which means that there is no common structure to share anymore (cf. Akasaka (1991) for a similar argument). The set of actors representing several readings of one lexical item can proceed concurrently, thus introducing further concurrency.

## 5 EVENT NETWORK SPECIFICATION OF A SAMPLE PARSE

We will now consider a partial event network in order to illustrate the parse of "*Compaq entwickelt einen Notebook mit einer 120-MByte-Harddisk*"[5]. At some point after reading the sentence, the configuration shown on Fig.3 will have been reached. The preposition [mit][6] is not yet integrated due to a mandatory valency that must be satisfied prior to making conceptual restrictions available. Upon establishment of a corresponding dependency between [mit] and [Harddisk] (Fig.3), [mit] starts to search for its head. This search results in the dependency tree depicted on Fig.4.

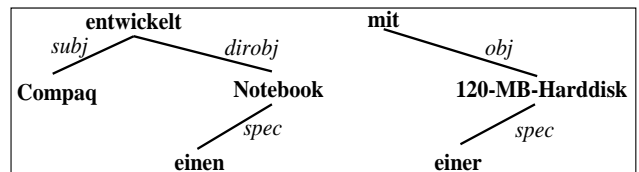

**Figure 3.** Configuration before application of "mit" via **searchHead**

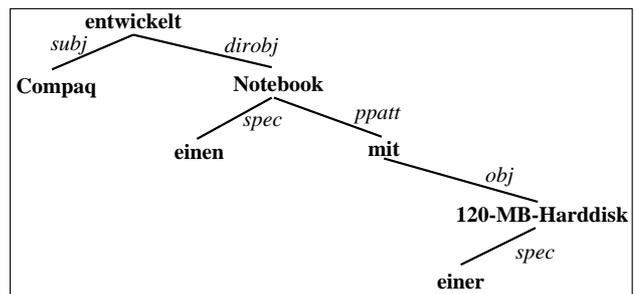

**Figure 4.** After establishment of dependency

The events caused by the satisfaction of the mandatory valency at [mit] (**headAccepted** event at top left of Fig.5)

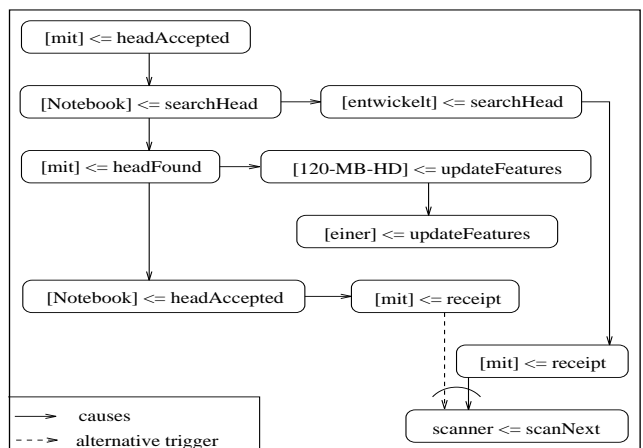

**Figure 5.** Event network

---

[5] A rough English translation of this reads as "*Compaq develops a notebook with a 120-MByte hard disk*". Notice that from a syntactic perspective either the verb "*entwickelt*" or the noun "*Notebook*" may take a prepositional phrase with "*mit*" specifying an instrument or a part, resp. This potential structural ambiguity does not occur in our model due to parallel evaluation of constraints in different knowledge sources.

[6] Word actors representing a lexical item *"x"* will be written as [x].

are specified in the event network in Fig.5. The dotted line indicates an alternative possibility how the **scanNext** event could have been triggered. Of the two receipt events, the last one taking place triggers the **scanNext** event (note that both involve the same actor, [mit], so that they must be ordered, even in a distributed system without global time).

# 6 CONCLUSIONS

The *ParseTalk* model of natural language understanding aims at the integration of a lexically distributed, dependency-based grammar specification with a solid formal foundation for concurrent, object-oriented parsing. The associated concurrent computation model is based on the actor paradigm of object-oriented programming, with several extensions relating to special requirements of natural language processing. These cover mechanisms for complex message distribution, synchronization in terms of request-reply protocols, and the distinction of distribution and computation events. We have shown how the semantic specification of actor systems can be used for the consideration of global interrelations of word actors at the grammar level (event type networks) and the parser level (event networks). While event type networks provide a general, global view on the behavioral aspects of our grammar specification, the current formalism still lacks the ability to support formal reasoning about computational properties of distributed systems, such as deadlock freeness, termination. On the other hand, event networks illustrate the computations during real parses, but do not allow predictions in general cases. Providing a type discipline for actor definitions may be a reasonable approach to fill the methodological gap between both layers of description.

The *ParseTalk* model has been experimentally validated by a prototype system, a parser for German. The current full-form lexicon contains a hierarchy of 54 word-class specifications and nearly 1000 lexical entries; a module for morphological analysis is under development. The parser's coverage is currently restricted to the analysis of assertional sentences, with focus on complex noun and prepositional phrases. The *ParseTalk* system is implemented in Smalltalk, with extensions that allow for coarse-grained parallelism through physical distribution in a workstation cluster (Xu, 1993) and asynchronous message passing. It is loosely coupled with the LOOM knowledge representation system (MacGregor & Bates, 1987). We currently use a knowledge base with 120 concept definitions covering the domain of information technology. Furthermore, an interactive graphical grammar/parser engineering workbench is supplied which supports the development and maintenance of the *ParseTalk* grammar system.

**Acknowledgments**

The work reported in this paper is funded by grants from DFG (grants no. Ha 2097/1-1, Ha 2097/1-2) within a special research programme on cognitive linguistics. We like to thank our colleagues, P. Neuhaus, M. Klenner, and Th. Hanneforth, for valuable comments and support.